# Aspect and Opinion Terms Extraction Using Double Embeddings and Attention Mechanism for Indonesian Hotel Reviews

Jordhy Fernando
School of Electrical Engineering and Informatics
Institut Teknologi Bandung
Bandung, Indonesia
jordhy.fernando@gmail.com

Masayu Leylia Khodra
School of Electrical Engineering and Informatics
Institut Teknologi Bandung
Bandung, Indonesia
masayu@stei.itb.ac.id

Ali Akbar Septiandri
AiryRooms
Jakarta, Indonesia
ali.septiandri@airy.com

*Abstract*—Aspect and opinion terms extraction from review texts is one of the key tasks in aspect-based sentiment analysis. In order to extract aspect and opinion terms for Indonesian hotel reviews, we adapt double embeddings feature and attention mechanism that outperform the best system at SemEval 2015 and 2016. We conduct experiments using 4000 reviews to find the best configuration and show the influences of double embeddings and attention mechanism toward model performance. Using 1000 reviews for evaluation, we achieved F1-measure of 0.914 and 0.90 for aspect and opinion terms extraction in token and entity (term) level respectively.

*Keywords—aspect and opinion terms extraction; attention mechanism; double embeddings*

## I. INTRODUCTION

Customer satisfaction is one of the key components contributing to the success of a business. Aspect-based sentiment analysis can be used to measure customer satisfaction. Business owner can use the result of this aspect-based sentiment analysis to determine aspects of their products or services that need to be improved.

One of the important tasks in aspect-based sentiment analysis is aspect and opinion terms extraction which aims to extract aspect terms and opinion terms from opinion texts [1]. An aspect term is a word or a phrase that describes an entity's attribute or feature that is the target of an opinion. An opinion term is a word or a phrase that shows subjective emotion toward an attribute or feature of an entity. For example, in a hotel review *"Tempat tidur di hotel ini tidak bersih"* (The bed in this hotel is not clean), extraction process returns *"Tempat tidur"* (bed) as aspect term and *"tidak bersih"* (not clean) as opinion term.

Aspect and/or opinion terms extraction research has been conducted by Wang et al. [2] and Xu et al. [3] that outperformed the best systems in the aspect-based sentiment analysis task on the International Workshop on Semantic Evaluation (SemEval) for aspect and opinion terms extraction.

Wang et al. [2] proposed a deep learning model for aspect and opinion terms extraction, named Coupled Multi-Layer Attentions (CMLA), with word embedding as its feature. The model is a multi-layer attention network, where each layer consists of a couple of attentions with tensor xoperators, one attention for aspect term extraction and the other for opinion term extraction. The model achieved F1-measure of 0.7073 and 0.7368 for aspect and opinion term extraction respectively using SemEval 2015 task 12 subtask 1 restaurant dataset [4].

Xu et al. [3] proposed a Convolutional Neural Network (CNN) model employing two types of pre-trained word embeddings, general-purpose embeddings and domain-specific embeddings, for aspect term extraction. The two embeddings are concatenated into one word embedding called double embeddings. The model achieved F1-measure of 0.7437 for aspect term extraction using SemEval 2016 task 5 subtask 1 restaurant dataset [5].

Wang et al. [2] and Xu et al. [3] approaches have not been applied for Indonesian reviews. This paper aims to perform aspect and opinion terms extraction in Indonesian hotel reviews by adapting CMLA architecture [2] and double embeddings mechanism [3]. The adaption in this paper is conducted by changing the English resources used in word embedding into Indonesian version.

The rest of this paper is structured as follows. Section 2 discusses related work. Section 3 describes the proposed approach. Section 4 discusses the results and analysis of the conducted experiment and evaluation. Finally, Section 5 concludes our research.

## II. RELATED WORK

Aspect and opinion terms extraction can be viewed as an information extraction task. One of the approaches used in aspect and opinion terms extraction is supervised learning. In supervised learning, aspect and opinion terms extraction is treated as a sequence labelling problem [6]. Jin and Ho [7] use Hidden Markov Model (HMM) with part of speech (POS) and lexical features to extract aspect and opinion terms. Jakob and Gurevych [8] used Conditional Random Field (CRF) with token, POS, short dependency path, word distance, and opinion sentence as it's features.

For Indonesian reviews, aspect and/or opinion terms extraction have been conducted by [9], [10], and [11] for restaurant domain as one of task in aspect-based sentiment analysis. Gojali and Khodra [9] performed aspect and opinion terms extraction by using CRF classifier with token and POS tag as features. Ekawati and Khodra [10] and Cahyadi and Khodra [11] only performed aspect term extraction. Ekawati and Khodra [10] used CRF classifier with distributional semantic model, lexical, and syntactic features. Cahyadi and Khodra [11] also used CRF classifier to do aspect term extraction. The features used in [11] are lexical features and output probabilities from Bidirectional Long Short-Term Memory (B-LSTM).

Recently, deep learning approaches have been proposed to extract aspect and/or opinion terms. Wang et al. [2] used attention mechanism [12] to identify the possibility of each token being an aspect or opinion term. The coupled multi-layer attentions that was proposed by [2] models the relations among tokens automatically without any syntactic/dependency parsing or linguistic resources as additional information for the input and achieves good performance for aspect and opinion terms extraction. The coupled attentions are used to exploit the correlations between aspect and opinion terms using tensor operators [2].

Xu et al. [3] use double embeddings that leverage both general embeddings and domain embeddings as a feature for a CNN model and let the CNN model decide which embeddings have more useful information. The experiment conducted in [3] demonstrated that double embedding mechanism achieved better performance for aspect terms extraction compared to the use of general embeddings or domain embeddings alone.

### III. Proposed Approach

As stated previously, the goal of this work is to extract aspect and opinion terms in Indonesian hotel reviews by adapting CMLA architecture [2] and double embeddings mechanism [3]. The architecture of the model used in this work can be seen in Fig. 1.

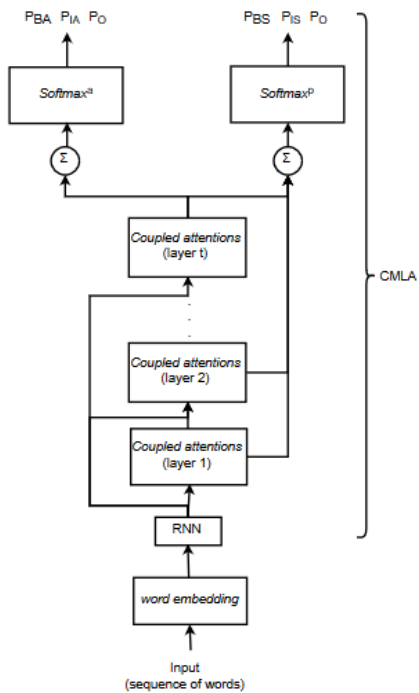

Fig. 1. Architecture of the model

Review texts are preprocessed to be used in training extraction model by using InaNLP [13]. The preprocess consists of sentence normalization, casefolding, and tokenization. Normalization is done because there are many informal words, abbreviations, and typos in the reviews. The normalization is conducted by transforming those words into formal words in Indonesian. Casefolding is performed so that each word will only have one representation the word embedding. For example, the review "*Kamar mndi mampet gak ada handuk dn sabun.*" would be normalized and casefolded into "*kamar mandi mampat tidak ada handuk dan sabun.*" (the bathroom is clogged there is no towel and soap.) The review is then tokenized into a list of tokens: ["*kamar*" (room), "*mandi*" (bath), "*mampat*" (clogged), "*tidak*" (no), "*ada*" (there is), "*handuk*" (towel), "*dan*" (and), "*sabun*" (soap), "*.*"].

Aspect and opinion terms extraction is treated as a sequence labelling problem with the BIO notation for the labels. There are five labels defined for a token: B-ASPECT (beginning of aspect term), I-ASPECT (inside of aspect term), B-SENTIMENT (beginning of opinion term), I-SENTIMENT (inside of sentiment term) and O (others). A single aspect term is a sequence with B-ASPECT at the beginning followed by I-ASPECT. A single opinion term is a sequence with B-SENTIMENT at the beginning followed by I-SENTIMENT. As an example, the review "*tempat tidur di hotel ini tidak bersih*" (The bed in this hotel is not clean) would be labelled "*tempat<B-ASPECT> tidur<I-ASPECT> di<O> hotel<O> ini<O> tidak<B-SENTIMENT> bersih <I-SENTIMENT>*" (The<B-ASPECT> bed<I-ASPECT> in<O> this<O> hotel<O> is<O> not<B-SENTIMENT> clean<I-SENTIMENT>).

We use CMLA model that is proposed by [2] to predict each token's label in a sequence. The architecture of the CMLA used in this work is as written in [2]. Table I shows the hyperparameters of model. In the experiment, we will try other variations of Recurrent Neural Network (RNN) to replace the Gated Recurrent Unit (GRU) used in CMLA. Specifically, we conduct experiment using GRU, LSTM, B-GRU, and B-LSTM and choose the one that gives the best performance based on the experiment as the final model. We implement and train the model using Keras [14].

TABLE I. CMLA Hyperparameters

| Hyperparameters | Description |
|---|---|
| Hidden units | The number of hidden units used by the model |
| Coupled attention layers | The number of coupled attention layers |
| Tensors dimension | The first dimension of the tensors used in the tensor operator |
| Dropout rate | The rate of the dropout used to regularize the model |

We use various types of word embeddings adapted from [3]. Specifically, we conduct experiment using double embeddings, general embeddings, domain embeddings, and hybrid embeddings as the feature used by the model and choose the word embedding that gives the best performance as the feature used by the final model. The description of each type of word embeddings can be seen in Table II.

We employed Indonesian Wikipedia articles obtained from Wikimedia [15] to train the general embeddings model. We have to preprocess the articles first because the articles obtained from Wikimedia are in XML format and contain information other than the articles' content, like the creator of the article and the title of the article. We only take the content of the article and remove all XML tags and other unnecessary symbols like symbols used to link to other Wikipedia article. There are 358052 articles used to train the general embeddings. For the training of domain embeddings model, we use Indonesian hotel reviews provided by AiryRooms. The reviews are preprocessed the same way as the preprocess done to the reviews used for training the aspect and opinion terms extraction model (except tokenization). We use 142810 reviews to train the domain embeddings. The hybrid embeddings use the combined corpus between Indonesian Wikipedia articles and Indonesian hotel reviews.

TABLE II. TYPES OF WORD EMBEDDINGS

| Word Embeddings | Description |
|---|---|
| Double embeddings | Concatenation of general embeddings and domain embeddings |
| General embeddings | Word embedding trained from general corpus |
| Domain embeddings | Word embedding trained from domain corpus |
| Hybrid embeddings | Word embedding trained from the combined corpus between general corpus and domain corpus |

All of the word embeddings are trained using fastText [16]. For the general embeddings and domain embeddings, we use the same dimension and number of iterations as in [3]. The embedding dimensions and number of iterations used to train the word embeddings can be seen in Table III. For the rest of the hyperparameters, we use the defaults in fastText. We use fastText for the word embedding because it can use subword N-gram embedding to calculate out-of-vocabulary word embeddings.

TABLE III. WORD EMBEDDINGS TRAINING PARAMETER

| Word Embeddings | Dimension | Iteration |
|---|---|---|
| General embeddings | 300 | 5 |
| Domain embeddings | 100 | 30 |
| Hybrid embeddings | 300 | 5 |

## IV. EXPERIMENT AND EVALUATION

We conduct experiments using 5000 Indonesian hotel reviews with a total of 78.604 tokens obtained from AiryRooms. We split the data into 3000 reviews for train data, 1000 reviews for validation data, and 1000 reviews for test data. The label distribution for each data can be seen in Table IV.

TABLE IV. LABEL DISTRIBUTION

| Label | Train data | Validation data | Test data |
|---|---|---|---|
| B-ASPECT | 5203 | 1802 | 1758 |
| I-ASPECT | 1709 | 583 | 584 |
| B-SENTIMENT | 7171 | 2475 | 2384 |
| I-SENTIMENT | 3179 | 1086 | 1067 |
| O | 29923 | 9974 | 9706 |
| **Total** | **47185** | **15920** | **15499** |

### A. Experiment Scenario

There are four experiment scenarios that we conducted in this work. The aim of each scenarios can be seen in Table V. The experiments are carried out in sequence starting with experiment P1, with each experiment uses the result from the previous experiment. We train the model using nadam optimizer with batch size of 32 categorical cross entropy as its loss function. We use early stopping with the patience set to 5 and the number of epochs set to 200. For experiment P1 and P2, we use the best hyperparameter values from [2]'s experiment. We use double embeddings as feature for experiment P1.

TABLE V. EXPERIMENT SCENARIOS ON VALIDATION DATA

| Experiment Id | Aim |
|---|---|
| P1 | Find the best variation of RNN to be used in the model |
| P2 | Find the best type of word embeddings to be used as feature |
| P3 | Find the best hyperparameters of the model |
| P4 | Compare the performance between model with attention mechanism and model without attention mechanism |

### B. Experiment Result

Table VI shows the result for experiment P1. There are four variations of RNN that we try: GRU, LSTM, B-GRU, and B-LSTM. The experiment result shows that the variation of RNN that gives the best performance for both token level and entity level is B-LSTM.

TABLE VI. EXPERIMENT P1 RESULTS

| RNN | Token Level | | | Entity Level | | |
|---|---|---|---|---|---|---|
| | P | R | F1 | P | R | F1 |
| GRU | 0.902 | 0.891 | 0.897 | 0.87 | 0.88 | 0.88 |
| LSTM | 0.900 | 0.903 | 0.902 | 0.87 | 0.89 | 0.88 |
| B-GRU | 0.887 | 0.919 | 0.902 | 0.87 | 0.91 | 0.89 |
| B-LSTM | **0.918** | **0.919** | **0.918** | **0.90** | **0.92** | **0.91** |

The result for experiment P2 can be seen in Table VII. We try four types of word embeddings: double embeddings, general embeddings, domain embeddings, and hybrid embeddings. The word embeddings that produce the best performance for token and entity level based on the experiment is double embeddings and the worst feature is hybrid embeddings that is trained from the combined corpus between general corpus and domain corpus.

There are several aspect and opinion terms that can be extracted by the model trained using double embeddings but cannot be extracted by the other models. For example, in the review *"kamar lumayan tetapi tolong dengan sangat resepsionis jangan judes dan galak begitu, tidak ada sopan santun. tidak pantas jadi resepsionis. semoga bisa diperbaiki"* (the room is ok but please for the receptionist don't be mean and fierce, there is no manners. not worthy of being a receptionist. hope it can be fixed), the models trained using double embeddings can extract the opinion term *"judes"* (mean), but the other models failed to extract that opinon term.

The model trained using double embeddings can extract several aspect and opinion terms that the model with domain embeddings failed to extract but can be extracted by the model with general embeddings. For example, in the review *"hotel tua, dapat kamar yang kuncinya rusak, kemudian dipindah ke kamar lain, ketel air rusak dan bau, kamar mandi tergenang air"* (old hotel, got the room with damaged key, then moved to another room, the kettle is broken and smelly, the bathroom is flooded), the models trained using double embeddings and general embeddings can extract the aspect term *"ketel air"* (kettle), but the model with domain embeddings failed to extract that aspect term.

The model trained using double embeddings can also extract several aspect and opinion terms that the model with general embeddings failed to extract but can be extracted by the model with domain embeddings. For example, in the review *"lantai kamar seperti ada lengketlengketnya, mungkin karena belum dipel"* (the room floor looks sticky, maybe because it hasn't been mopped), the models trained using double embeddings and domain embeddings can extract the aspect term *"lantai kamar"* (the room floor), but the model with general embeddings failed to extract that aspect term.

TABLE VII. EXPERIMENT P2 RESULTS

| Word embeddings | Token Level | | | Entity Level | | |
|---|---|---|---|---|---|---|
| | P | R | F1 | P | R | F1 |
| Double embeddings | **0.918** | **0.919** | **0.918** | **0.90** | **0.92** | **0.91** |
| General embeddings | 0.893 | 0.904 | 0.899 | 0.87 | 0.89 | 0.88 |
| Domain embeddings | 0.904 | 0.913 | 0.911 | 0.89 | 0.91 | 0.90 |
| Hybrid embeddings | 0.887 | 0.898 | 0.892 | 0.86 | 0.89 | 0.87 |

In order to find the best hyperparameters configuration of the model, we investigate multiple values on the number of hidden units, number of coupled attention layer, tensors dimension, and dropout rate. The number of hyperparameters that we tried in total is 81 combinations. The three best experiment results for hyperparameters configuration for token and entity level can be seen in Table VIII and Table IX. The best hyperparameters values are 50 for number of hidden units, 2 for number of coupled attention layer, 20 for tensors dimension, and 0.5 for dropout rate. Based on the experiment result, the three best hyperparameters configurations give nearly the same performance.

After we found the best hyperparameters configuration, we conduct experiment P4 to see the effect of attention mechanism (coupled attention) by comparing the performance between model with attention mechanism (CMLA) and model without attention mechanism (B-LSTM + Softmax). Table X shows the result for experiment P4.

Based on the experiment result, the use of attention mechanism improves the performance for aspect and opinion terms extraction. There are several aspect and opinion terms that can be extracted by CMLA but failed to be extracted by the model without attention mechanism. For example, in the review *"acnya tidak dingin dan tidak ada tisunya sama laundrynya kebetulan trouble jadi handuknya cuma dapat 1"* (the air conditioner is not cold and there is no tissue also coincidentally the laundry has a problem so I only got 1 towel), CMLA can extract the aspect terms *"tisunya"* (the tissue) and *"handuknya"* (the towel) by exploiting the relations between aspect terms and opinion terms (in this case, the corresponding opinion terms for the two aspect terms are *"tidak ada"* (there is no) and *"cuma dapat 1"* (only got 1)). The model without mechanism can only extract the opinion terms and failed to extract the aspect terms.

TABLE VIII. EXPERIMENT P3 TOP RESULTS FOR TOKEN LEVEL

| Hyperparameter | | Score | | |
|---|---|---|---|---|
| Name | Value | Precision | Recall | F1 |
| Hidden units | 50 | **0.918** | **0.919** | **0.918** |
| Tensors dim | 20 | | | |
| Dropout rate | 0.5 | | | |
| Coupled attention layer | 2 | | | |
| Hidden units | 75 | 0.908 | 0.921 | 0.915 |
| Tensors dim | 15 | | | |
| Dropout rate | 0.5 | | | |
| Coupled attention layer | 1 | | | |
| Hidden units | 75 | 0.893 | 0.917 | 0.913 |
| Tensors dim | 20 | | | |
| Dropout rate | 0.2 | | | |
| Coupled attention layer | 1 | | | |

TABLE IX. EXPERIMENT P3 TOP RESULTS FOR ENTITY LEVEL

| Hyperparameter | | Score | | |
|---|---|---|---|---|
| Name | Value | Precision | Recall | F1 |
| Hidden units | 50 | **0.90** | **0.92** | **0.91** |
| Tensors dim | 20 | | | |
| Dropout rate | 0.5 | | | |
| Coupled attention layer | 2 | | | |
| Hidden units | 75 | 0.90 | 0.91 | 0.91 |
| Tensors dim | 15 | | | |
| Dropout rate | 0.5 | | | |
| Coupled attention layer | 1 | | | |
| Hidden units | 75 | 0.90 | 0.90 | 0.90 |
| Tensors dim | 20 | | | |
| Dropout rate | 0.2 | | | |
| Coupled attention layer | 1 | | | |

TABLE X. EXPERIMENT P4 RESULTS

| Model | Token Level | | | Entity Level | | |
|---|---|---|---|---|---|---|
| | P | R | F1 | P | R | F1 |
| CMLA | **0.918** | **0.919** | **0.918** | **0.90** | **0.92** | **0.91** |
| B-LSTM + Softmax | 0.903 | 0.904 | 0.903 | 0.87 | 0.90 | 0.88 |

C. Evaluation Result

The model trained using the best variation of RNN, type of word embeddings, and hyperparameters configuration is evaluated by using test data that consists of 1000 reviews. We compare the performance of our trained model with a baseline model. We use B-LSTM-CRF that is proposed by [19] and implemented using anaGo [20] as the baseline model. The baseline model achieves good results for Named Entity Recognition (NER) in four languages. We use the default configurations provided by anaGo to train the baseline model. We use early stopping with the patience set to 5 and the number of epochs set to 200. Table XI and Table XII show the evaluation result for token and entity level. FKA represents the evaluation result of the model built in this work and BS represents the evaluation result of the baseline model. The evaluation results show that the model in this work outperforms the baseline model.

One of the cases where misclassifications occur is when the aspect or opinion term failed to be normalized. An example for this case is in the review *"tmpatnya nyamn. brsh. dan ramah."* (the place is comfortable. clean. and friendly.) where the model fails to extract the opinion term *"brsh"* (clean) that should be normalized into *"bersih"* (clean). However, the opinion term *"nyamn"* (comfortable) can be extracted by them model even though it is not normalized into *"nyaman"* (comfortable). This happens because the word *"nyamn"* and *"nyaman"* have a cosine similarity of 0.717 which indicates that those two words are similar, while the word *"brsh"* and *"bersih"* have a cosine similarity of 0.084 indicating that the two words are different.

Another case of misclassification occurs in the review *"bagus, ramah, siklus udara di toilet perlu diperbaiki."* (nice, friendly, the air cycle in the toilet needs to be repaired). The model fails to extract the aspect term *"siklus udara"* (air cycle) because the aspect term *"siklus udara"* (air cycle) never appear in the training data.

TABLE XI. EVALUATION RESULTS FOR TOKEN LEVEL

| Label | Precision | | Recall | | F1 | |
|---|---|---|---|---|---|---|
| | FKA | BS | FKA | BS | FKA | BS |
| B-ASPECT | 0.913 | 0.879 | 0.919 | 0.887 | 0.916 | 0.883 |
| I-ASPECT | 0.842 | 0.837 | 0.906 | 0.793 | 0.873 | 0.814 |
| B-SENTIMENT | 0.939 | 0.927 | 0.939 | 0.909 | 0.939 | 0.918 |
| I-SENTIMENT | 0.907 | 0.849 | 0.865 | 0.823 | 0.886 | 0.836 |
| O | 0.957 | 0.936 | 0.957 | 0.945 | 0.957 | 0.940 |
| Average | **0.912** | 0.886 | **0.917** | 0.871 | **0.914** | 0.878 |

TABLE XII. EVALUATION RESULTS FOR ENTITY LEVEL

| Label | Precision | | Recall | | F1 | |
|---|---|---|---|---|---|---|
| | FKA | BS | FKA | BS | FKA | BS |
| ASPECT | 0.88 | 0.84 | 0.91 | 0.85 | 0.89 | 0.85 |
| SENTIMENT | 0.91 | 0.89 | 0.91 | 0.88 | 0.91 | 0.89 |
| Average | **0.895** | 0.865 | **0.91** | 0.865 | **0.90** | 0.87 |

Misclassification also occurs when there are aspect or opinion terms that are not about hotel's attributes. For example, in the review *"lokasi strategis. dekat pasar tradisional untuk cari sarapan murah meriah. dekat juga dengan bakpia patuk 25. indomaret. untuk complimentary snak agar diberikan lagi jika menginap lebih dari satu malam."* (strategic location. near the traditional market to find cheap breakfast. close to bakpia patuk 25. indomaret. for complimentary snacks to be given again if staying more than one night), the word *"sarapan"* (breakfast) and *"murah meriah"* (cheap) should not be extracted as aspect and opinion terms because those words are not about the breakfast provided by hotel.

Another case of misclassification occurs in the review *"sering mati listrik jika pemanas air dan ac dinyalakan."* (power outages often happen if the water heater and air conditioner are turned on). The model extracts the word *"ac"* (air conditioner) as an aspect term whereas it should not be extracted as an aspect term because it does not have any opinion term director toward it.

The last case of misclassification occurs when there is subjectivity in the annotation. An example for this case can be found the review *"tempatnya tenang. hanya parkiran tidak nyaman. swimming pool kurang untuk kids."* (the place is calm. parking is uncomfortable. swimming pool is not good enough for kids.). The model extracted the word *"pool"* as an aspect term which causes it to be counted as false prediction because the true aspect term is *"swimming pool"*. This is subjective because both *"pool"* and *"swimming pool"* can be considered as aspect term.

V. CONCLUSION

We adapted the architecture of CMLA and double embeddings mechanism to do aspect and opinion terms extraction in Indonesian hotel reviews. The adaption is conducted by investigating various types of RNN for the CMLA and various types of word embeddings to be used as the feature for the model. We also employ Indonesian resources to train the word embedding. The experiment results demonstrated that double embeddings and the use of attention mechanism improves the performance of aspect and opinion terms extraction. We achieved F1-measure of 0.914 dan 0.90 in aspect and opinion terms extraction of Indonesian hotel review using the model with best configurations for token and entity level respectively, while the baseline model achieved F1-measure of 0.878 and 0.87 for token and entity level respectively.

The performance of the model can be improved by adding more training data. The quality of the data can be improved by increasing the number of annotators and reannotation because the annotation of the data is still imperfect. The performance of the model can also be improved by improving the preprocess step, especially the normalization, because there are still some words that are not normalized. Besides fastText, other word embeddings models like GloVe and word2vec should be taken into consideration in the experiment to improve performance. Increasing the training data to train the word embeddings which can also improve performance by improving the word representation.

ACKNOWLEDGMENT

We would like to thank AiryRooms for providing Indonesian hotel reviews used in this research to build the aspect and opinion terms extraction model.

REFERENCES

[1] B. Pang and L. Lee. 2008. Opinion mining and sentiment analysis. *Foundations and Trends in Information Retrieval* 2(1-2).

[2] W. Wang, S. J. Pan, D. Dahlmeier, and X. Xiao. 2017. Coupled multi-layer attentions for co-extraction of aspect and opinion terms. In

*Thirty-First AAAI Conference on Artificial Intelligence*, pp. 3316–3322.

[3] H. Xu, B. Liu, L. Shu, and P. S. Yu. 2018. Double embeddings and cnn-based sequence labeling for aspect extraction. In *Proceedings of the 56th Annual Meeting of the Association for Computational Linguistics*. Association for Computational Linguistics.

[4] M. Pontiki, D. Galanis H. Papageorgiou, S. Manandhar, and I. Androutsopoulos. 2015. Semeval-2015 task 12: Aspect based sentiment analysis. In *SemEval*, pp. 486–495.

[5] M. Pontiki, et al. 2016. Semeval-2016 task 5: Aspect based sentiment analysis. In *SemEval*, pp. 19-30.

[6] B. Liu. 2012. *Sentiment Analysis and Opinion Mining*. Morgan & Claypool Publishers.

[7] W. Jin and H. H. Ho. 2009. A novel lexicalized hmm-based learning framework for web opinion mining. In *ICML*, 465–472.

[8] N. Jakob and I. Gurevych. 2010. Extracting opinion targets in a single and cross-domain setting with conditional random fields. In *Proceedings of the 2010 Conference on Empirical Methods in Natural Language Processing*.

[9] S. Gojali and M. L. Khodra. 2016. Aspect based sentiment analysis for review rating prediction. In *ICAICTA*.

[10] D. Ekawati and M. L. Khodra. 2017. Aspect-based sentiment analysis for indonesian restaurant reviews. In *ICAICTA*.

[11] A. Cahyadi and M. L. Khodra. 2018. Aspect-based sentiment analysis using convolutional neural network and bidirectional long short-term memory. In *ICAICTA*.

[12] D. Bahdanau, K. Cho, and Y. Bengio. 2014. Neural machine translation by jointly learning to align and translate. In *CoRR abs/1409.0473*.

[13] A. Purwarianti, A. Andhika, A. F. Wicaksono, I. Afif and F. Ferdian. 2016. InaNLP: Indonesia Natural Language Processing Toolkit. In *ICAICTA*.

[14] F. Chollet, "Keras," 2015. [Online]. Available: https://keras.io.

[15] "Wikimedia," [Online]. Available: https://www.wikimedia.com/.

[16] P; Bojanowski, E. Grave, A. Joulin, and T. Mikolov. 2016. Enriching word vectors with subword information. *arXiv preprint arXiv:1607.04606*.

[17] F. Pedregosa, et al. 2011. Scikit-learn: Machine Learning in Python. In *Journal of Machine Learning Research*, vol. 12, pp. 2825-2830

[18] https://github.com/chakki-works/seqeval.

[19] G. Lample, M. Ballesteros, S. Subramanian, K. Kawakami and C. Dyer. 2016. Neural architectures for named entity recognition. *arXiv preprint arXiv:1603.01360*.

[20] anaGo. 2018. "Bidirectional LSTM-CRF for Sequence Labeling," [Online]. Available: https://github.com/Hironsan/anago